%% file: main.tex
\documentclass[runningheads]{llncs}

\usepackage[mobile]{eccv}

\usepackage{eccvabbrv}

\usepackage{booktabs}

\usepackage[accsupp]{axessibility}  
\usepackage{graphicx}
\usepackage{tikz}
\usepackage{comment}
\usepackage{amsmath,amssymb} 
\usepackage{color}

\usepackage{ulem} 

\usepackage{booktabs}
\usepackage{colortbl}
\usepackage{pifont}
\newcommand{\xmark}{\ding{55}}
\usepackage{gensymb}
 \usepackage{adjustbox}
\usepackage{wrapfig}
\usepackage{caption}
\usepackage{comment}
\usepackage{cancel}
\usepackage{rotating} %
\usepackage[colorlinks=true, allcolors=blue]{hyperref}

\usepackage{multirow}

\usepackage{orcidlink}

\begin{document}

\title{3DEgo: 3D Editing on the Go!}

\titlerunning{3DEgo: 3D Editing on the Go!}

\author{Umar Khalid\inst{1,*}\orcidlink{0000-0002-3357-9720}\and Hasan Iqbal\inst{2,*}\orcidlink{0009-0005-2162-3367} \and Azib Farooq\inst{3}\orcidlink{0009-0006-7867-2546}
  \and Jing Hua\inst{2}\orcidlink{0000-0002-3981-2933}  \and Chen Chen\inst{1}\orcidlink{0000-0003-3957-7061} 
}

\authorrunning{U.~Khalid et al.}

\institute{University of Central Florida, Orlando, FL, USA \and
Department of Computer Science, Wayne State University, Detroit, MI, USA \and Department of Computer Science and Software Engineering, Miami University, Oxford, OH, USA}

\maketitle

{ \renewcommand{\thefootnote}%
    {\fnsymbol{footnote}}
  \footnotetext[1]{Equal Contribution}}

\begin{abstract}
We introduce \textbf{3DEgo} to address a novel problem of directly synthesizing photorealistic 3D scenes from monocular videos guided by textual prompts. Conventional methods construct a text-conditioned 3D scene through a three-stage process, involving pose estimation using Structure-from-Motion (SfM) libraries like COLMAP, initializing the 3D model with unedited images, and iteratively updating the dataset with edited images to achieve a 3D scene with text fidelity. Our framework streamlines the conventional multi-stage 3D editing process into a single-stage workflow by overcoming the reliance on COLMAP and eliminating the cost of model initialization. We apply a diffusion model to edit video frames prior to 3D scene creation by incorporating our designed \textit{noise blender module} for enhancing multi-view editing consistency, a step that does not require additional training or fine-tuning of T2I diffusion models.  \textbf{3DEgo} utilizes 3D Gaussian Splatting to create 3D scenes from the multi-view consistent edited frames, capitalizing on the inherent temporal continuity and explicit point cloud data. 3DEgo demonstrates remarkable editing precision, speed, and adaptability across a variety of video sources, as validated by extensive evaluations on six datasets, including our own prepared GS25 dataset. Project Page: \url{https://3dego.github.io/}

\end{abstract}

\input{sec/02_intro}

\input{sec/03_related_work}
\input{sec/04_method}
\input{sec/05_evaluation}

\input{sec/06_limitations}

\input{sec/06_conclusion}

\section*{Acknowledgement}
This work was partially supported by the NSF under Grant Numbers OAC-1910469 and OAC-2311245.
\bibliographystyle{splncs04}
\bibliography{egbib}
\end{document}

%% file: sec/02_intro.tex
\section{Introduction}

\label{sec:intro}

In the pursuit of constructing photo-realistic 3D scenes from monocular video sources, it is a common practice to use the Structure-from-Motion (SfM) library, COLMAP~\cite{schonberger2016structure} for camera pose estimation. This step is critical for aligning frames extracted from the video, thereby facilitating the subsequent process of 3D scene reconstruction. To further edit these constructed 3D scenes, a meticulous process of frame-by-frame editing based on textual prompts is often employed~\cite{zhuang2023dreameditor}. Recent works, such as IN2N~\cite{haque2023instruct}, estimate poses from frames using  SfM~\cite{schonberger2016structure} to initially train an unedited 3D scene.  Upon initializing a 3D model, the training dataset is iteratively updated by adding edited images at a consistent rate of editing.   This process of iterative dataset update demands significant computational resources and time. Due to challenges with initial edit consistency, IN2N\cite{haque2023instruct} training necessitates the continuous addition of edited images to the dataset over a significantly large number of iterations. This issue stems from the inherent limitations present in Text-to-Image (T2I) diffusion models\cite{brooks2023instructpix2pix, rombach2022high}, where achieving prompt-consistent edits across multiple images—especially those capturing the same scene—proves to be a formidable task~\cite{kim2023collaborative, dong2024vica}. Such inconsistencies significantly undermine the effectiveness of 3D scene modifications, particularly when these altered frames are leveraged to generate unique views. 
\begin{wrapfigure}{r}{0.57\textwidth} 
    \centering
    \includegraphics[width=0.55\textwidth, trim={4.2cm 5.3cm 4cm 3cm}, clip]{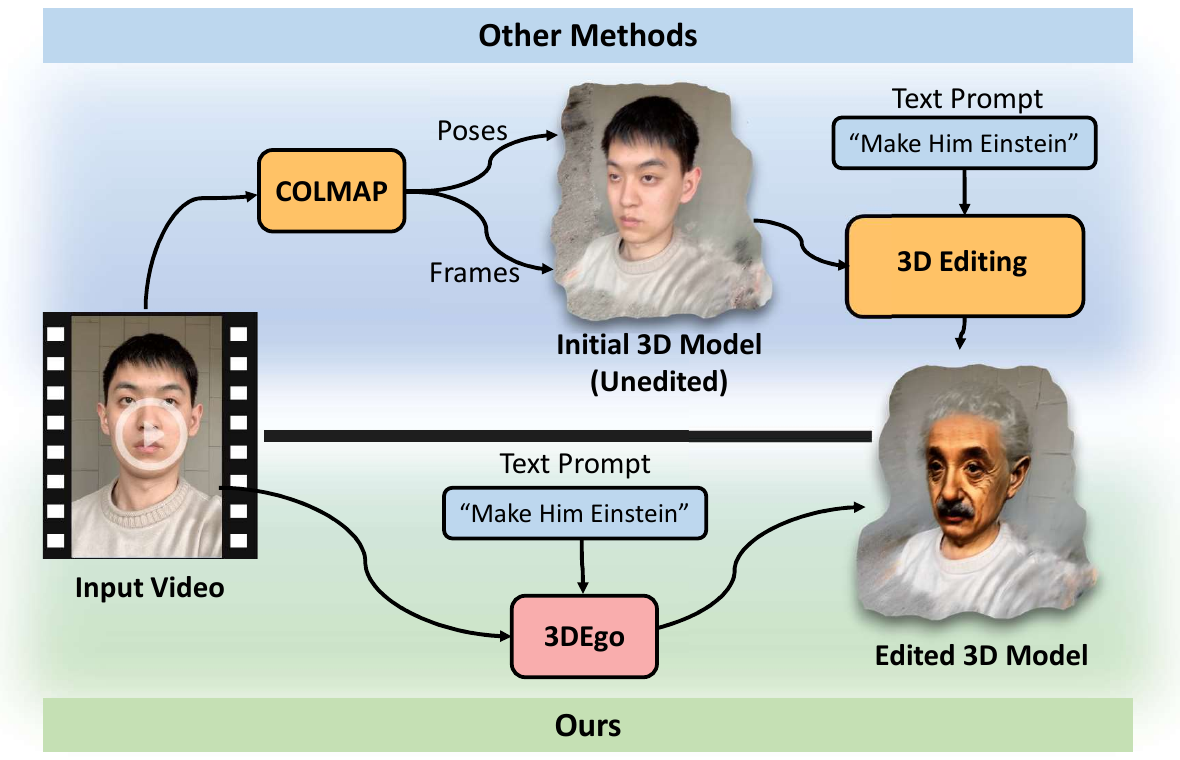} 
    \caption{\footnotesize{Our method, \textbf{3DEgo}, streamlines the 3D editing process by merging a three-stage workflow into a singular, comprehensive framework. This efficiency is achieved by bypassing the need for COLMAP~\cite{schonberger2016structure} for pose initialization and avoiding the initialization of the model with unedited images, unlike other existing approaches~\cite{haque2023instruct,dong2024vica,kim2023collaborative}.}}
    \label{fig:method_diff}
\end{wrapfigure}

In this work, we address a\textit{ novel problem} of efficiently reconstructing 3D scenes directly from monocular videos without using COLMAP~\cite{schonberger2016structure} aligned with the editing textual prompt. Specifically, we apply a diffusion model~\cite{brooks2023instructpix2pix} to edit every frame of a given monocular video before creating a 3D scene. To address the challenge of consistent editing across all the frames, we introduce a novel \textit{noise blender module}, which ensures each new edited view is conditioned upon its adjacent, previously edited views. This is achieved by calculating a weighted average of image-conditional noise estimations such that closer frames exert greater influence on the editing outcome. Our editing strategy utilizes the IP2P~\cite{brooks2023instructpix2pix} 2D editing diffusion model, which effectively employs both conditional and unconditional noise prediction. Consequently, our method achieves multi-view consistency without the necessity for extra training or fine-tuning, unlike prior approaches~\cite{dong2024vica, long2024wonder3d,weng2023consistent123}.  For 3D scene synthesis based on the edited views, our framework utilizes the Gaussian Splatting (GS)~\cite{kerbl20233d} technique, capitalizing on the temporal continuity of video data and the explicit representation of point clouds. Originally designed to work with pre-computed camera poses, 3D Gaussian Splatting presents us with the possibility to synthesize views and construct edited 3D scenes from monocular videos without the need for SfM pre-processing, overcoming one of NeRF's significant limitations~\cite{lin2021barf}.

\begin{figure}[t!]

\centering
    \includegraphics[width=0.85\linewidth, trim={2cm 2.5cm 2cm 1cm}, clip]{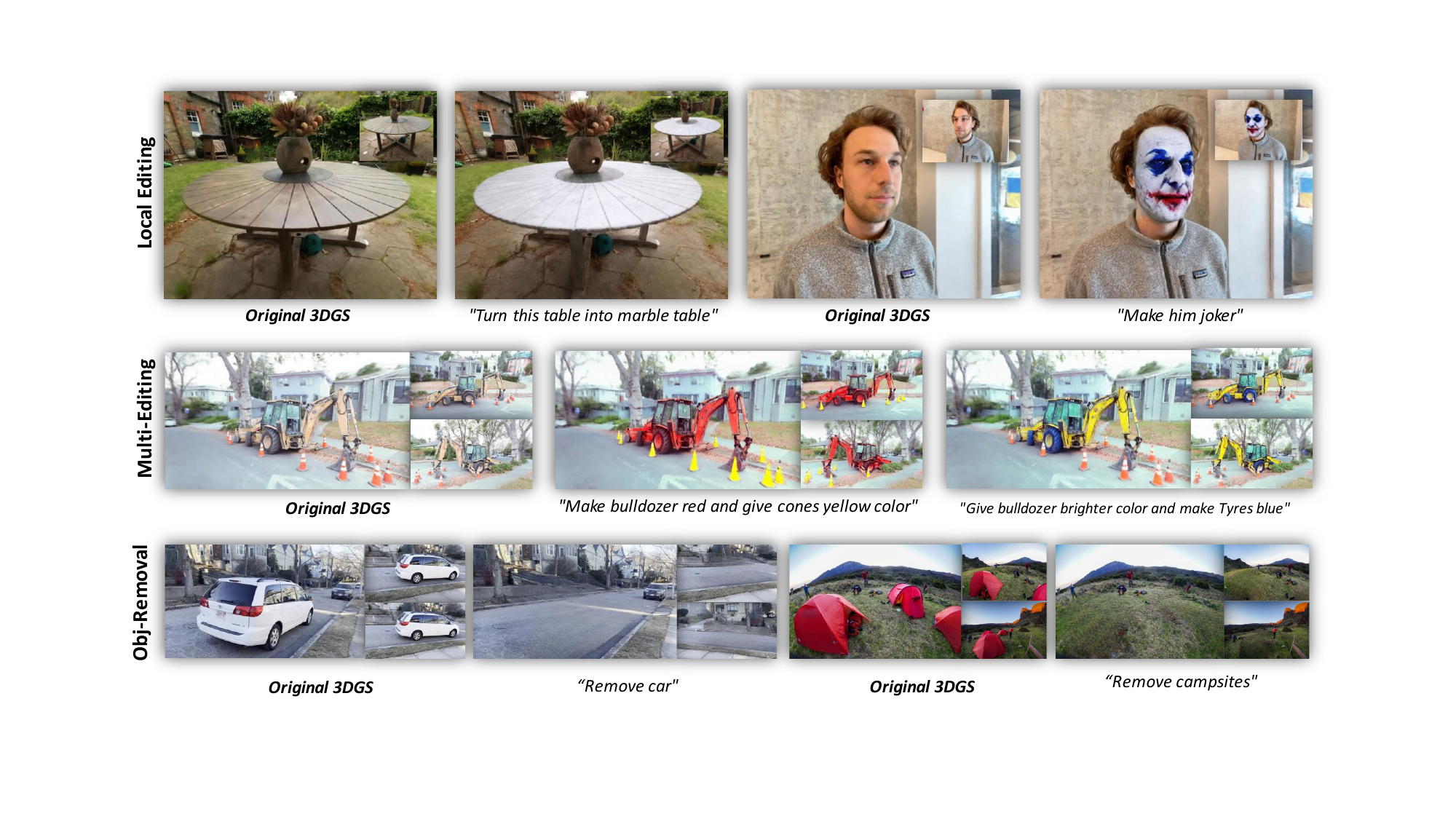}
\caption{\footnotesize{\textbf{3DEgo} offers rapid, accurate, and adaptable 3D editing, bypassing the need for original 3D scene initialization and COLMAP poses. This ensures compatibility with videos from any source, including casual smartphone captures like the \textbf{Van} 360-degree scene. The above results identify three cases challenging for IN2N~\cite{haque2023instruct}}, where our method can convert a monocular video into customized 3D scenes using a streamlined, single-stage reconstruction process. }
\label{fig:figteaser}

\end{figure}

Our method grows the 3D Gaussians of the scene continuously, from the edited frames, as the camera moves, eliminating the need for pre-computed camera poses and 3D model initialization on original un-edited frames to identify an affine transformation that maps the 3D Gaussians from frame $i$ to accurately render the pixels in frame $i+1$. Hence, our method \textbf{3DEgo} condenses a three-stage 3D editing process into a single-stage, unified and efficient framework as shown in Figure~\ref{fig:method_diff}.
Our contributions are as follows:
\begin{itemize}
    \item We tackle the novel challenge of directly transforming monocular videos into 3D scenes guided by editing text prompts, circumventing conventional 3D editing pipelines.
    \item We introduce a unique auto-regressive editing technique that enhances multi-view consistency across edited views, seamlessly integrating with pre-trained diffusion models without the need for additional fine-tuning.
    \item We propose a COLMAP-free method using 3D Gaussian splatting for reconstructing 3D scenes from casually captured videos. This technique leverages the video's continuous time sequence for pose estimation and scene development, bypassing traditional SfM dependencies.
     \item We present an advanced technique for converting 2D masks into 3D space, enhancing editing accuracy through Pyramidal Gaussian Scoring (PGS), ensuring more stable and detailed refinement.
    \item Through extensive evaluations on six datasets—including our custom \textbf{GS25} and others like IN2N, Mip-NeRF, NeRFstudio Dataset, Tanks \& Temples, and CO3D-V2—we demonstrate our method's enhanced editing precision and efficiency, particularly with \textit{360-degree} and \textit{casually recorded} videos, as illustrated in Fig.~\ref{fig:figteaser}.
\end{itemize}

%% file: sec/03_related_work.tex
\section{Related Work}
\label{sec:related_work}

A growing body of research is exploring diffusion models for text-driven image editing, introducing techniques that allow for precise modifications based on user-provided instructions~\cite{nichol2021glide, ramesh2022hierarchical, rombach2022high, saharia2022photorealistic}. While some approaches require explicit before-and-after captions~\cite{hertz2022prompt} or specialized training~\cite{ruiz2023dreambooth}, making them less accessible to non-experts, IP2P~\cite{brooks2023instructpix2pix} simplifies the process by enabling direct textual edits on images, making advanced editing tools more widely accessible.

Recently, diffusion models have also been employed for 3D editing, focusing on altering the geometry and appearance of 3D scenes~\cite{michel2022text2mesh, hong2022avatarclip, wang2022clip, kobayashi2022decomposing, tschernezki2022neural, bao2023sine, gao2023textdeformer, noguchi2021neural, liu2022nerf, li2022climatenerf, xu2022deforming, yang2022neumesh, brooks2023instructpix2pix, li2023focaldreamer, khalid2023latenteditor, karim2023free}.

Traditional NeRF representations, however, pose significant challenges for precise editing due to their implicit nature, leading to difficulties in localizing edits within a scene. Earlier efforts have mainly achieved global transformations~\cite{wang2023nerf, chiang2022stylizing, huang2022stylizednerf, nguyen2022snerf, zhang2022arf, wu2022palettenerf}, with object-centric editing remaining a challenge. IN2N~\cite{haque2023instruct} introduced user-friendly text-based editing, though it might affect the entire scene. Recent studies~\cite{zhuang2023dreameditor,dong2024vica,kim2023collaborative} have attempted to tackle local editing and multi-view consistency challenges within the IN2N framework~\cite{haque2023instruct}. Yet, no existing approaches in the literature offer pose-free capabilities, nor can they create a text-conditioned 3D scene from arbitrary video footage. Nevertheless, existing 3D editing methods~\cite{haque2023instruct,zhuang2023dreameditor} universally necessitate Structure-from-Motion (SfM) preprocessing. Recent studies like Nope-NeRF~\cite{bian2023nope}, BARF~\cite{lin2021barf}, and SC-NeRF~\cite{jeong2021self} have introduced methodologies for pose optimization and calibration concurrent with the training of \textit{(unedited)} NeRF. 

In this study, we present a novel method for constructing 3D scenes directly from textual prompts, utilizing monocular video frames without dependence on COLMAP poses~\cite{schonberger2016structure}, thus addressing unique challenges. Given the complexities NeRF's implicit nature introduces to simultaneous 3D reconstruction and camera registration, our approach leverages the advanced capabilities of 3D Gaussian Splatting (3DGS)~\cite{kerbl20233d}  alongside a pre-trained 2D editing diffusion model for efficient 3D model creation.

%% file: sec/04_method.tex
\section{Method}

\label{sec:method}
Given a sequence of unposed images alongside camera intrinsics, we aim to recover the camera poses in sync with the edited frames and reconstruct a photo-realistic 3D scene conditioned on the textual prompt. 
\subsection{Preliminaries}
\label{sec:gau_splat}

In the domain of 3D scene modeling, 3D Gaussian splatting~\cite{kerbl20233d} emerges as a notable method. The method's strength lies in its succinct Gaussian representation coupled with an effective differential rendering technique, facilitating real-time, high-fidelity visualization. This approach models a 3D environment using a collection of point-based 3D Gaussians, denoted as $\mathcal{H}$  where each Gaussian $h = \{\mu, \Sigma, c, \alpha\}$. Here, $\mu \in \mathbb{R}^3$ specifies the Gaussian's center location, $\Sigma\in \mathbb{R}^{3 \times 3}$ is the covariance matrix capturing the Gaussian's shape, $c \in \mathbb{R}^3$ is the color vector in RGB format represented in
the three degrees of spherical harmonics (SH) coefficients, and $\alpha \in \mathbb{R}$ denotes the Gaussian's opacity level.
To optimize the parameters of 3D Gaussians to represent the scene, we need to render them into images in a differentiable manner. The rendering is achieved by approximating the projection of a 3D Gaussian along the depth dimension into pixel coordinates expressed as:

\begin{equation}
    C = \sum_{p \in \mathcal{P}} c_p \tau_p \prod_{k=1}^{p-1} (1 - \alpha_k),
    \label{eq:1}
\end{equation}
where $\mathcal{P}$ are ordered points overlapping the pixel, and $\tau_p = \alpha_p e^{-\frac{1}{2}(x_p)^T\Sigma^{-1}(x_p)}$ quantifies the Gaussian's contribution to a specific image pixel, with $x_p$ measuring the distance from the pixel to the center of the $p$-th Gaussian. 
In the original 3DGS, initial Gaussian parameters are refined to fit the scene, guided by\textit{ ground truth poses }obtained using SfM. Through differential rendering, the Gaussians' parameters, including position $\mu$, shape $\Sigma$, color $c$, and opacity $\alpha$, are adjusted using a photometric loss function.

\begin{wrapfigure}{r}{0.45\textwidth} 
    \centering
    
   \includegraphics[width=0.45\textwidth, clip]{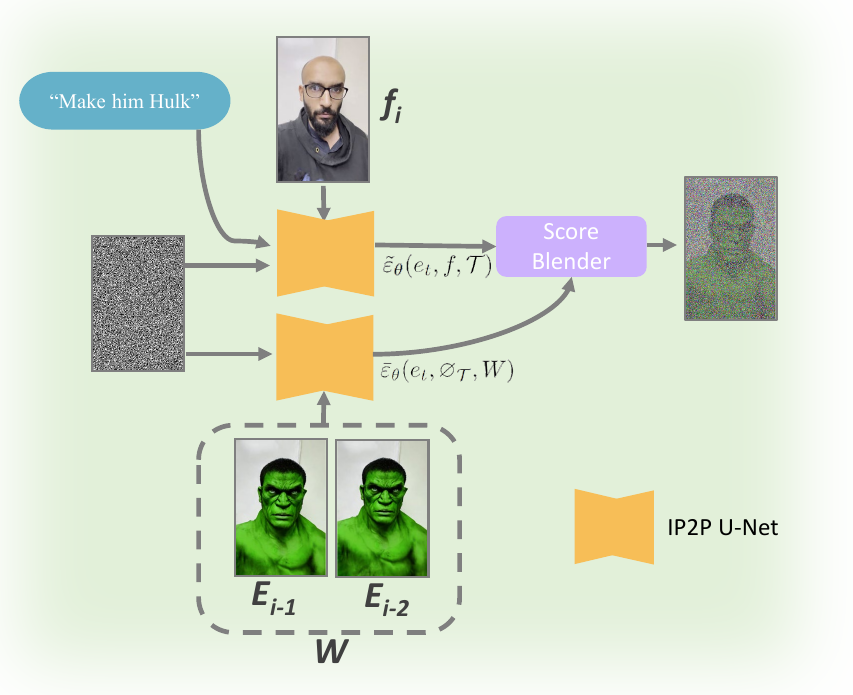}
\caption{\footnotesize{\textbf{Autoregressive Editing.} At each denoising step, the model predicts $w+1$ separate noises, which are then unified via weighted noise blender (Eq.~\ref{noise_blender}) to predict ${\varepsilon}_\theta (e_t, f, \mathcal{T}, W)$. }}
\label{fig:fig6}

\end{wrapfigure}

\subsection{Multi-View Consistent 2D Editing} 

In the first step, we perform 2D editing with key editing areas (KEA) based on the user-provided video, $V$, and editing prompt, $\mathcal{T}$.

\label{KEA}

From the given video $V$, we extract frames $\{f_1, f_2, \ldots, f_N\}$. Analyzing the textual prompt $\mathcal{T}$ with a Large Language Model $\mathcal{L}$ identifies key editing attributes $\{A_1, A_2, \ldots, A_k\}$, essential for editing, expressed as $\mathcal{L}(\mathcal{T}) \rightarrow \{A_1, A_2, \ldots, A_k\}$. Utilizing these attributes, a segmentation model $\mathcal{S}$ delineates editing regions in each frame $f_i$ by generating a mask $M_i$, with KEA marked as 1, and others as 0. The segmentation operation is defined as, $  \mathcal{S}(f_i, \{A_1, A_2, \ldots, A_k\}) \rightarrow M_i, \quad \forall i \in \{1, \ldots, N\}.$ Subsequently, a 2D diffusion model $\mathcal{E}$ selectively edits these regions in $f_i$, as defined by $M_i$, producing edited frames $\{E_1, E_2, \ldots, E_N\}$ under guidance from $\mathcal{T}$, such that $\mathcal{E}(f_i, M_i) \rightarrow E_i.$

\subsubsection{Consistent Multi-View2D Editing.}
\label{2dedtitng}
As discussed above, differing from IN2N \cite{haque2023instruct} that incorporates edited images gradually over several training iterations, our approach involves editing the entire dataset at once before the training starts. We desire 1) each edited frame, $E_i$
follows the editing prompt, $\mathcal{T}$, 2) retain the original images' semantic content, and 3) the edited images, $\{E_1, E_2, \ldots, E_N\}$ are consistent with each other. 

\noindent\textbf{\textit{(i)} Multi-view Consistent Mask.} As $\mathcal{S}$ doesn't guarantee consistent masks across the views of a casually recorded monocular video,  we utilize a zero-shot point tracker~\cite{rajivc2023segment} to ensure uniform mask generation across the views. The procedure starts by identifying query points in the initial video frame using the ground truth mask. Query points are extracted from these ground truth masks employing the K-Medoids~\cite{park2009simple} sampling method. This method utilizes the cluster centers from K-Medoids clustering as query points. This approach guarantees comprehensive coverage of the object's various sections and enhances resilience to noise and outliers. 

\noindent\textbf{\textit{(ii)}Autoregressive Editing.} To address the issue of preserving consistency across multiple views, we employ an autoregressive method that edits frames in sequence, with IP2P~\cite{brooks2023instructpix2pix} editing restricted to the Key Editing Areas (KEA) as delineated by the relevant masks. Instead of editing each frame independently from just the input images - a process that can vary significantly between adjacent images - we integrate an \textit{autoregressive} editing technique where the frame to be edited is conditioned on already edited adjacent frames. 


As discussed above, we incorporate IP2P~\cite{brooks2023instructpix2pix} as a 2D editing diffusion model. The standard noise prediction from IP2P's backbone that includes both conditional and unconditional editing is given as, 
\begin{equation}
\scalebox{0.73}{
$ \tilde{\varepsilon}_\theta (e_t, f, \mathcal{T}) = \varepsilon_\theta(e_t, \varnothing_f, \varnothing_{\mathcal{T}}) + s_f \big(\varepsilon_\theta(e_t,  f, \varnothing_{\mathcal{T}}) - \varepsilon_\theta(e_t,  \varnothing_f, \varnothing_{\mathcal{T}})\big) + s_{\mathcal{T}} \big(\varepsilon_\theta(e_t, f, \mathcal{T}) - \varepsilon_\theta(e_t,  f, \varnothing_{\mathcal{T}})\big)$
}
\label{eq:ip2p}
\end{equation}

where $ s_f$ and $s_{\mathcal{T}}$ are image and textual prompt guidance scale.We suggest enhancing the noise estimation process with our autoregressive training framework. Consider a set of \(w\) views, represented as \(W=\{E_n\}_{n=1}^{w}\). Our goal is to model the distribution of the \(i\)-th view image by utilizing its \(w\) adjacent, already edited views. To achieve this, we calculate image-conditional noise estimation, $\varepsilon_\theta(e_t,  E, \varnothing_{\mathcal{T}})$ across all frames in \(W\). The equation to compute the weighted average \(\bar{\varepsilon}_\theta\) of the noise estimates from all edited frames within \(W\), employing \(\beta\) as the weight for each frame, is delineated as follows:
\begin{equation}
\bar{\varepsilon}_\theta(e_t, \varnothing_{\mathcal{T}},W) = \sum_{n=1}^{w} \beta_n \varepsilon^n_\theta (e_t, E_n, \varnothing_{\mathcal{T}})
\label{eq:colab}
\end{equation}

Here, \(E_n\) represents the \(n\)-th edited frame within \(W\), and \(\beta_n\) is the weight assigned to the \(n\)-th frame's noise estimate. The condition that the sum of all \(\beta\) values over \(w\) frames equals 1 is given by as, $\sum_{n=1}^{w} \beta_n = 1$. This ensures that the weighted average is normalized. As we perform 2D editing without any pose priors, our weight parameter $\beta$ is independent of the angle offset between the frame to be edited, $f_n$ and already edited frames in $W$. To assign weight parameters with exponential decay, ensuring the closest frame receives the highest weight, we can use an exponential decay function for the weight \(\beta_n\) of the \(n\)-th frame in $W$. By employing a decay factor \(\lambda_d\) (0 < \(\lambda_d\) < 1), the weight of each frame decreases exponentially as its distance from the target frame increases. The weight \(\beta_n\) for the \(n\)-th frame  is defined as, $
    \beta_n = \lambda_d^{w-n}$. 
 This ensures the, $E$ closest to the target, $f$ (\(n = 1\)) receives the highest weight. To ensure the sum of the weights to 1, each weight is normalized by dividing by the sum of all weights, $
\beta_n = \frac{\lambda^{w-n}}{\sum_{j=1}^{w} \lambda^{w-j}}
$.This normalization guarantees the sum of \(\beta_n\) across all \(n\) equals 1, adhering to the constraint \(\sum_{n=1}^{w} \beta_n = 1\). 

Our editing path is determined by the sequence of frames from the captured video. Therefore, during the editing of frame \(f_n\), we incorporate the previous \(w\) edited frames into the set \(W\), assigning the highest weight \(\beta\) to \(E_{n-1}\). Using Eq.~\ref{eq:ip2p} and Eq.~\ref{eq:colab}, we define our score estimation function as following:
\begin{equation}
\label{noise_blender}
    {\varepsilon}_\theta (e_t, f, \mathcal{T}, W)= \gamma_f \tilde{\varepsilon}_\theta (e_t, f, \mathcal{T})+\gamma_E\bar{\varepsilon}_\theta(e_t, \varnothing_{\mathcal{T}},W)
\end{equation}
where \(\gamma_f\) is a hyperparameter that determines the influence of the original frame undergoing editing on the noise estimation, and \(\gamma_E\) represents the significance of the noise estimation from adjacent edited views.

\subsection{3D Scene Reconstruction}

After multi-view consistent 2D editing is achieved across all frames of the given video, $V$, we leverage the edited frames $E_i$ and their corresponding masks $M_i$ to construct a 3D scene without any SfM pose initialization. Due to the explicit nature of 3DGS~\cite{kerbl20233d},  determining the camera poses is essentially equivalent to estimating the transformation of a collection of 3D Gaussian points. Next, we will begin by introducing an extra Gaussian parameter for precise local editing. Subsequently, we will explore relative pose estimation through incremental frame inclusion. Lastly, we will examine the scene expansion, alongside a discussion on the losses integrated into our global optimization strategy.

\subsubsection{3D Gaussians Parameterization for Precise Editing.}
\label{sec:gaussian_parameter}
Projecting KEA (see Section~\ref{KEA}) into 3D Gaussians, $\mathcal{H}$, using $M$ for KEA identity assignment, is essential for accurate editing. Therefore, we introduce a vector, $m$ associated with the Gaussian point, $h = \{\mu, \Sigma, c, \alpha, m\}$ in the 3D Gaussian set, $\mathcal{H}_i$ of the $i_{th}$ frame. The parameter  $m$ is a learnable vector of length 2 corresponding to the number of labels in the segmentation map, $M$. We optimize the newly introduced parameter $m$  to represent KEA identity during training. However, unlike the view-dependent Gaussian parameters, the KEA Identity remains uniform across different rendering views. Gaussian KEA identity ensures the continuous monitoring of each Gaussian's categorization as they evolve, thereby enabling the selective application of gradients, and the exclusive rendering of targeted objects, markedly enhancing processing efficiency in intricate scenes.

Next, we delve into the training pipeline inspired by ~\cite{fu2023colmapfree3dgaussiansplatting, bian2023nope} in detail which consists of two stages: \textit{(i)} Relative Pose Estimation, and \textit{(ii)} Global 3D Scene Expansion.

\subsubsection{Per Frame View Initialization.} \label{initialization}To begin the training process, , we randomly pick a specific frame, denoted as \(E_i\). We then employ a pre-trained monocular depth estimator, symbolized by \(\mathcal{D}\), to derive the depth map \(D_i\) for \(E_i\). Utilizing $D_i$, which provides strong geometric cues independent of camera parameters, we initialize 3DGS with points extracted from monocular depth through camera intrinsics and orthogonal projection. This initialization step involves learning a set of 3D Gaussians $\mathcal{H}_i$ to minimize the photometric discrepancy between the rendered and current frames $E_i$. The photometric loss, $\mathcal{L}_{rgb}$,  optimize the conventional 3D Gaussian parameters including color $c$, covariance $\Sigma$, mean $\mu$, and opacity $\alpha$. However, to initiate the KEA identity and adjust \(m_g\) for 3D Gaussians, merely relying on \(\mathcal{L}_{rgb}\) is insufficient. Hence, we propose the KEA loss, denoted as \(\mathcal{L}_{KEA}\), which encompasses the 2D mask \(M_i\) corresponding to $E_i$. We learn the KEA identity of each Gaussian point during training by applying $\mathcal{L}_{KEA}$ loss ($\mathcal{L}_{KEA}$). Overall, 3D Gaussian optimization is defined as,

\begin{equation}
    \mathcal{H}_i^* = \arg \min_{c, \Sigma, \mu, \alpha} \mathcal{L}_{rgb}(\mathcal{R} (\mathcal{H}_i), E_i)+ \arg \min_{m} \mathcal{L}_{KEA}(\mathcal{R} (\mathcal{H}_i), M_i),
    \label{eq:5}
\end{equation}
where $\mathcal{R}$ signifies the 3DGS rendering function. The photometric loss $\mathcal{L}_{rgb}$ as introduced in~\cite{kerbl20233d} is a blend of $\mathcal{L}_1$ and D-SSIM losses:
\begin{equation}
    \mathcal{L}_{rgb} = (1-\gamma) \mathcal{L}_1 + \gamma \mathcal{L}_{\text{D-SSIM}},
\end{equation}


$\mathcal{L}_{KEA}$ has two components to it. \textit{(i)} 2D Binary Cross-Entropy Loss, and  \textit{(ii)} 3D Jensen-Shannon Divergence (JSD) Loss, and is defined as,

\begin{equation}
    \mathcal{L}_{KEA} = \lambda_{BCE}\mathcal{L}_{BCE}+\lambda_{JSD}\mathcal{L}_{JSD}
\end{equation}

Let $\mathcal{N}$ be the total number of pixels in the $M$, and $\mathcal{X}$ represent the set of all pixels. We calculate binary cross-entropy loss $\mathcal{L}_{BCE}$ as following, 

\begin{equation}\scalebox{0.85}{$
\begin{aligned}
\mathcal{L}_{BCE} = & -\frac{1}{\mathcal{N}} \sum_{x \in \mathcal{X}} \bigg[ M_i(x) \log\left(\mathcal{R} (\mathcal{H}_i, m)(x)\right) 
& + (1 - M_i(x)) \log\left(1 - \mathcal{R} (\mathcal{H}_i, m)(x)\right) \bigg]
\end{aligned}$
}
\end{equation}
where $M(x)$ is the value of the ground truth mask at pixel $x$, indicating whether the pixel belongs to the foreground (1) or the background (0). The sum computes the total loss over all pixels, and the division by $\mathcal{N}$ normalizes the loss, making it independent of the image size. A rendering operation, denoted as $\mathcal{R}(\mathcal{H}_i, m)(x)$, produces $m_{\mathcal{R}}$ for a given pixel $x$, which represents the weighted sum of the vector $m$ values for the overlapping Gaussians associated with that pixel. Here, $m$ and $m_{\mathcal{R}}$ both have a dimensionality of 2 which is intentionally kept the same as the number of classes in mask labels. We apply \textit{softmax} function on $m_{\mathcal{R}}$ to extract KEA identity given as,  $
 \text{KEA Identity} = \text{softmax}(m_{\mathcal{R}})$. The \textit{softmax} output is interpreted as either 0, indicating a position outside the KEA, or 1, denoting a location within the KEA.

To enhance the accuracy of Gaussian KEA identity assignment, we also introduce an unsupervised 3D Regularization Loss to directly influence the learning of Identity vector \(m\). This 3D Regularization Loss utilizes spatial consistency in 3D, ensuring that the Identity vector, $m$ of the top \(k\)-nearest 3D Gaussians are similar in feature space. Specifically, we employ a symmetrical and bounded loss based on the Jensen-Shannon Divergence,

\begin{equation}
\scalebox{0.85}{ 
$\begin{aligned}
\mathcal{L}_{\text{JSD}} = \frac{1}{2YZ} \sum_{y=1}^{Y}\sum_{z=1}^{Z} \left[ S(m_y) \log\left(\frac{2S(m_y)}{S(m_y) + S(m'_z)}\right) + S(m'_z) \log\left(\frac{2S(m'_z)}{S(m_y) + S(m'_z)}\right) \right]
\end{aligned}$
}
\end{equation}
Here, $S$ indicates the \textit{softmax} function, and $m'_{z}$ represents the \textit{$z^{th}$} Identity vector from the $Z$ nearest neighbors in 3D space.

\noindent\textbf{Relative Pose Initialization.} Next, the relative camera pose is estimated for each new frame added to the training scheme. ${\mathcal{H}_i}^*$ is transformed via a learnable SE-3 affine transformation $\mathcal{M}_i$ to the subsequent frame $i+1$, where $\mathcal{H}_{i+1}= \mathcal{M}_i  \odot \mathcal{H}_i$. Optimizing transformation $\mathcal{M}_i$ entails minimizing the photometric loss between the rendered image and the next frame $E_{i+1}$,
\begin{equation}
    {\mathcal{M}_i}^* = \arg \min_{\mathcal{M}_i} \mathcal{L}_{rgb}(\mathcal{R}(\mathcal{M}_i \odot\mathcal{H}_i), E_{i+1}),
\end{equation}

In this optimization step, we keep the attributes of ${\mathcal{H}_i}^*$ fixed to distinguish camera motion from other Gaussian transformations such as pruning, densification, and self-rotation. 
Applying the above 3DGS initialization to sequential image pairs enables inferring relative poses across frames. However, accumulated pose errors could adversely affect the optimization of a global scene. To tackle this challenge, we propose the gradual, sequential expansion of the 3DGS.

\subsubsection{Gradual 3D Scene Expansion.}\label{method:joint} As illustrated above, beginning with frame $E_i$, we initiate with a collection of 3D Gaussian points, setting the camera pose to an orthogonal configuration. Then, we calculate the relative camera pose between frames $E_i$ and $E_{i+1}$. After estimating the relative camera poses, we propose to expand the 3DGS scene. This all-inclusive 3DGS optimization refines the collection of 3D Gaussian points, including all attributes, across $I$ iterations, taking the calculated relative pose and the two observed frames as inputs.  With the availability of the next frame $E_{i+2}$ after $I$ iterations, we repeat the above procedure: estimating the relative pose between $E_{i+1}$ and $E_{i+2}$, and expanding the scene with all-inclusive 3DGS.

To perform all-inclusive 3DGS optimization, we increase the density of the Gaussians currently under reconstruction as new frames are introduced. Following~\cite{kerbl20233d}, we identify candidates for densification by evaluating the average magnitude of position gradients in view-space. To focus densification on these yet-to-be-observed areas, we enhance the density of the universal 3DGS every $I$ step, synchronized with the rate of new frame addition. We continue to expand the 3D Gaussian points until the conclusion of the input sequence. Through the repetitive application of both frame-relative pose estimation and all-inclusive scene expansion, 3D Gaussians evolve from an initial partial point cloud to a complete point cloud that encapsulates the entire scene over the sequence. In our global optimization stage, we still utilize the $\mathcal{L}_{KEA}$ loss as new Gaussians are added during densification.

\noindent\textbf{Pyramidal Feature Scoring.} While our 2D consistent editing approach, detailed in Section~\ref{2dedtitng}, addresses various editing discrepancies, to rectify any residual inconsistencies in 2D editing, we introduce a pyramidal feature scoring method tailored for Gaussians in Key Editing Areas (KEA) identified with an identity of 1. This method begins by capturing the attributes of all Gaussians marked with KEA identity equal to 1 during initialization, establishing them as anchor points. With each densification step, these anchors are updated to mirror the present attributes of the Gaussians. Throughout the training phase, an intra-point cloud loss, $\mathcal{L}_{ipc}$ is utilized to compare the anchor state with the Gaussians' current state, maintaining that the Gaussians remain closely aligned with their initial anchors. $\mathcal{L}_{ipc}$ is defined as the weighted mean square error (MSE) between the anchor Gaussian and current Gaussian parameters with the older Gaussians getting higher weightage.

\noindent\textbf{Regularizing Estimated Pose.}
Further, to optimize the estimated relative pose between subsequent Gaussian set, we introduce point cloud loss, $\mathcal{L}_{pc}$  similar as in~\cite{bian2023nope}. While we expand the scene, $\mathcal{L}_{ipc}$ limits the deviation of the Gaussian parameters while $\mathcal{L}_{pc}$ regularizes the all-inclusive pose estimation. 
\begin{equation}
\mathcal{L}_{pc} =  D_{\text{Chamfer}} (\mathcal{M}_i^*\mathcal{H}_i^*, \mathcal{H}_{i+1}^*)
\end{equation}
Given two Gaussians, \(h_i\) and \(h_j\), each characterized by multiple parameters encapsulated in their parameter vectors \(\vec{\theta}_i\) and \(\vec{\theta}_j\) respectively, the Chamfer distance \(D_{\text{Chamfer}}\) between \(h_i\) and \(h_j\) can be formulated as:

\begin{equation}
D_{\text{Chamfer}}(h_i, h_j) = \sum_{p \in \vec{\theta}_i} \min_{q \in \vec{\theta}_j} \|p - q\|^2 + \sum_{q \in \vec{\theta}_j} \min_{p \in \vec{\theta}_i} \|q - p\|^2
\end{equation}

This equation calculates the Chamfer distance by summing the squared Euclidean distances from each parameter in \(h_i\) to its closest counterpart in \(h_j\), and vice versa, thereby quantifying the similarity between the two Gaussians across all included parameters such as color, opacity, etc. Combining all the loss components results in the total loss function during scene expansion,

\begin{equation}
    \mathcal{L}_{T} = \lambda_{rgb}\mathcal{L}_{rgb} + \lambda_{KEA}\mathcal{L}_{KEA} + \lambda_{ipc}\mathcal{L}_{ipc} + \lambda_{pc}\mathcal{L}_{pc}
\end{equation}
where \(\lambda_{rgb}\), \(\lambda_{KEA}\), \(\lambda_{ipc}\) and \(\lambda_{pc}\) act as weighting factors for the respective loss terms.

%% file: sec/05_evaluation.tex
\begin{figure}[t!]
\centering
    \includegraphics[width=0.85\linewidth, trim={0cm 1.5cm 0cm 0cm}, clip]{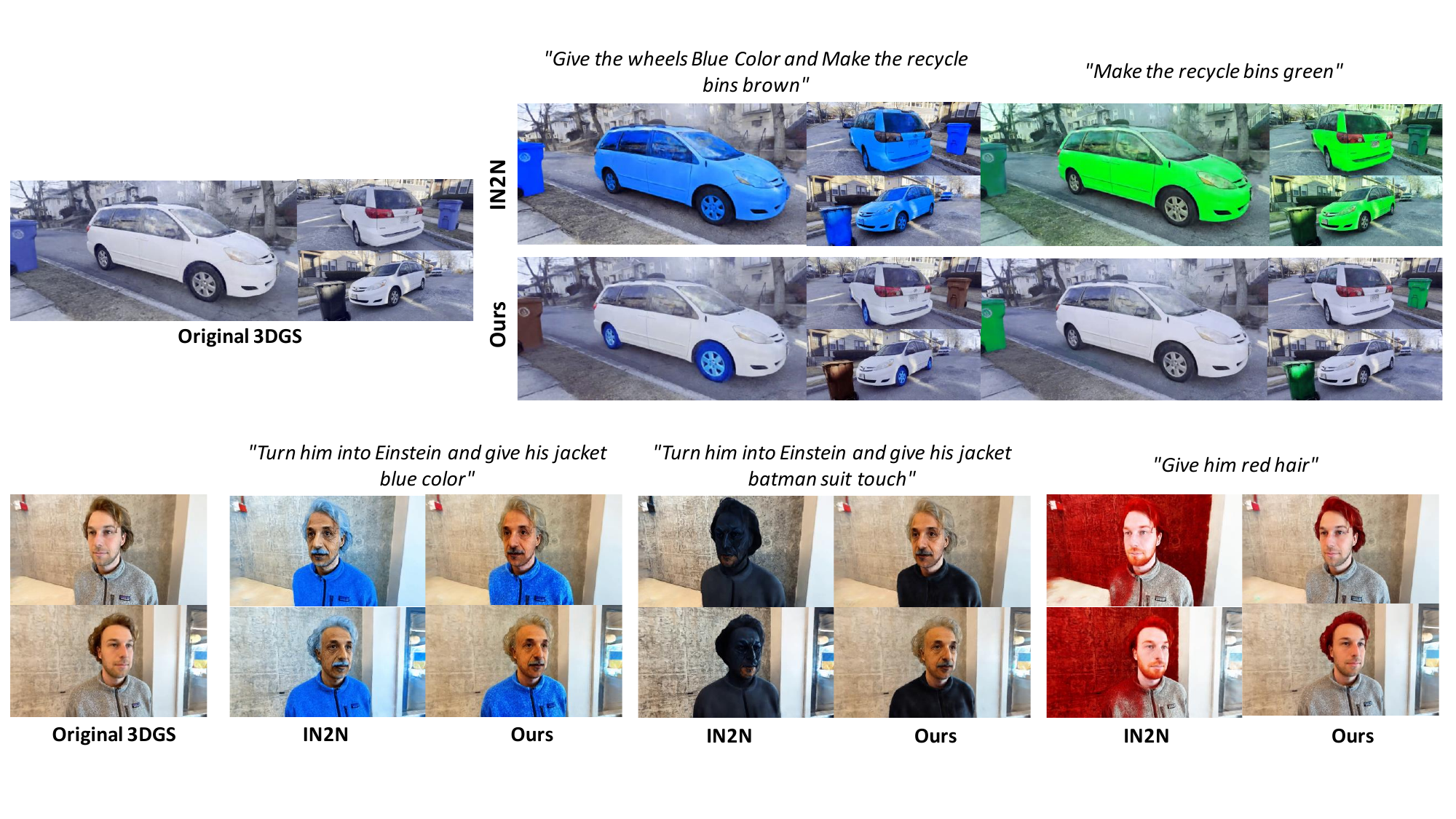}
\caption{\footnotesize {Qualitative comparison of our method with the IN2N~\cite{haque2023instruct} over two separate scenes.  When the editing prompt requests "Give the wheels Blue Color and Make the recyclebins brown," IN2N~\cite{haque2023instruct} inadvertently alters the complete van color to blue as well, instead of just changing the tire color. It must be noted that IN2N~\cite{haque2023instruct} uses poses from COLMAP, while \textit{3DEgo} estimates poses while constructing the 3D scene.}}
\label{fig:comparison1}
\end{figure}

\section{Evaluation}
\subsection{Implementation Details}
In our approach, we employ PyTorch~\cite{paszke2019pytorch} for the development, specifically focusing on 3D Gaussian splatting. GPT-3.5 Turbo~\cite{brown2020language} is used for identifying the editing attributes to identify the KEA. For segmentation purposes, SAM~\cite{kirillov2023segment} is used to generate the masks based on the key editing attributes identifying the KIA. For zero-shot point tracking, we employ a point-tracker as proposed in~\cite{rajivc2023segment}. The editing tasks are facilitated by the Instruct Pix2Pix~\cite{brooks2023instructpix2pix} 2D diffusion model by incorporating the masks to limit the editing within KEA. Additional details are in {\color{blue}{\textit{supplementary material}}}.

\subsection{Baseline and Datasets}

We carry out experiments across a variety of public datasets as well as our prepared \textbf{GS25} dataset. 

\begin{wraptable}{r}{0.55\textwidth}
\centering
\caption{\footnotesize Average runtime efficiency across 25 edits from the GS25 dataset (Approx. minutes).}
    \scalebox{0.65}{
    \begin{tabular}{l|c|c|c}
    \toprule
{Method} & COLMAP& Model Initialization & Scene Editing \\
    \midrule
    \textbf{Instruct-N2N}~\cite{haque2023instruct} & \cellcolor{red!25}13min& \cellcolor{red!25}22min & \cellcolor{blue!25}250min \\
    \textbf{Ours} & \cellcolor{green!25}\xmark & \cellcolor{green!25}\xmark & \cellcolor{blue!25}25min \\
    \bottomrule
    \end{tabular}}%
\label{tab:comparisontime}
\end{wraptable}

\noindent\textbf{GS25 Dataset} comprises 25 casually captured monocular videos using mobile phones for comprehensive 3D scene analysis.  This approach ensures the dataset's utility in exploring and enhancing 360-degree real-world scene reconstruction technologies. To further assess the efficacy of the proposed 3D editing framework, we also conducted comparisons across 5 public datasets: (i) IN2N~\cite{haque2023instruct}, (ii) Mip-NeRF~\cite{barron2022mip},(iii) NeRFstudio Dataset \cite{tancik2023nerfstudio}, (iv) Tanks \& Temples~\cite{Knapitsch2017} and (v) CO3D-V2~\cite{reizenstein2021common}.  We specifically validate the robustness of our approach on the CO3D dataset, which comprises thousands of object-centric videos. 
In our study, we introduce a unique problem, making direct comparisons with prior research challenging. Nonetheless, to assess the robustness of our method, we contrast it with state-of-the-art (SOTA) 3D editing techniques that rely on poses derived from COLMAP. Additionally, we present quantitative evaluations alongside pose-free 3D reconstruction approaches, specifically NoPeNeRF~\cite{bian2023nope}, and BARF~\cite{lin2021barf}. In the pose-free comparison, we substitute only our 3D scene reconstruction component with theirs while maintaining our original editing framework unchanged. We present a time-cost analysis in Table~\ref{tab:comparisontime} that underscores the rapid text-conditioned 3D reconstruction capabilities of 3DEgo.

\begin{figure}[t!]

\centering
    \includegraphics[width=0.85\linewidth, clip]{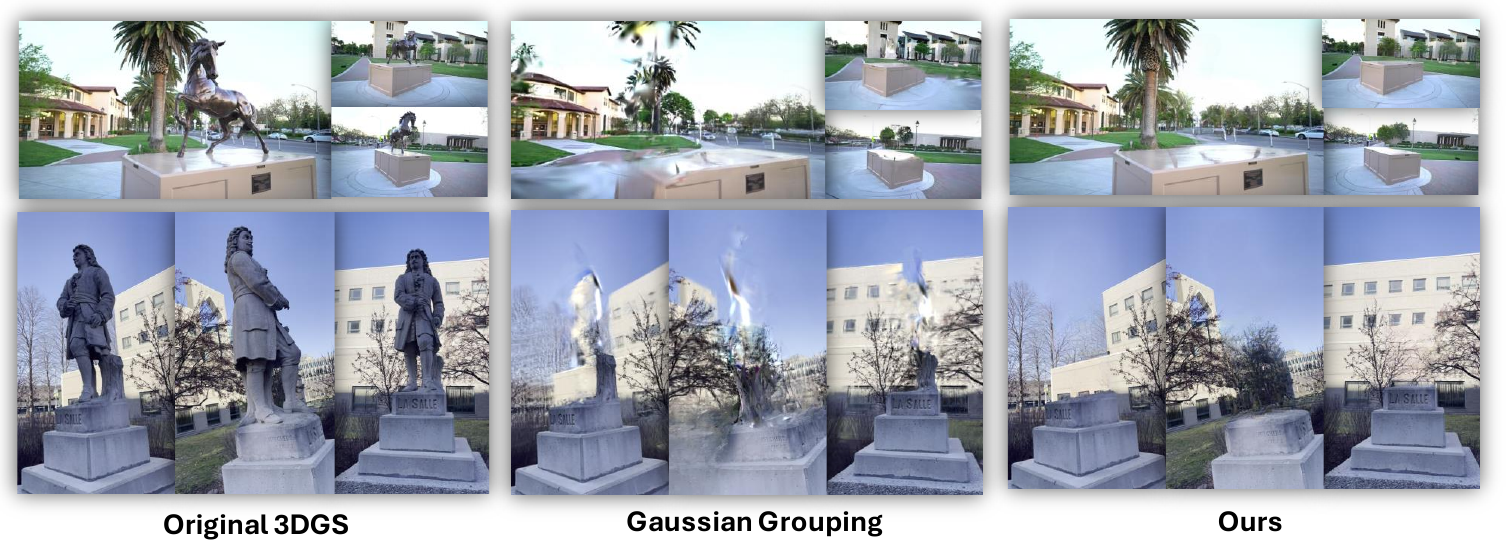}
\caption{\footnotesize{Our approach surpasses Gaussian Grouping~\cite{ye2023gaussian} in 3D object elimination across different scenes from GS25 and Tanks \& Temple datasets. \textit{3DEgo} is capable of eliminating substantial objects like statues from the entire scene while significantly minimizing artifacts and avoiding a blurred background.}}
\label{fig:comparison-removal}
\end{figure}
\begin{table}[b]
\centering
\footnotesize
\setlength{\tabcolsep}{2pt}
\renewcommand{\arraystretch}{1}
\caption{\footnotesize \textbf{Comparing With Pose-known Methods.} Quantitative evaluation of 200 edits across GS25, IN2N, Mip-NeRF, NeRFstudio, Tanks \& Temples, and CO3D-V2 datasets against the methods that incorporate COLMAP poses. The top-performing results are emphasized in bold.}
\resizebox{\textwidth}{!}{
\begin{tabular}{l|ccc|ccc|ccc}
\hline
\multirow{2}{*}{\textbf{Datasets}} & \multicolumn{3}{c|}{\textbf{DreamEditor}} & \multicolumn{3}{c|}{\textbf{IN2N}} & \multicolumn{3}{c}{\textbf{Ours}} \\
& CTIS$\uparrow$ & CDCR$\uparrow$ & E-PSNR$\uparrow$ & CTIS$\uparrow$ & CDCR$\uparrow$ & E-PSNR$\uparrow$ & CTIS$\uparrow$ & CDCR$\uparrow$ & E-PSNR$\uparrow$ \\
\hline
GS25 (Ours) &0.155&0.886&22.750&0.142&0.892&23.130&\textbf{0.169}&\textbf{0.925}&\textbf{23.660}\\
Mip-NeRF &0.149&0.896&23.920&0.164&\textbf{0.917}&22.170&\textbf{0.175}&0.901&\textbf{24.250}\\
NeRFstudio &0.156&0.903&23.670&\textbf{0.171}&0.909&\textbf{25.130}&0.163&\textbf{0.931}&24.990\\
CO3D-V2 &0.174&0.915&24.880&0.163&0.924&25.180&\textbf{0.179}&\textbf{0.936}&\textbf{26.020}\\
IN2N &0.167&0.921&24.780&0.179&0.910&\textbf{26.510}&\textbf{0.183}&\textbf{0.925}&26.390\\
Tanks \& Temples &0.150&0.896&23.970&\textbf{0.170}&0.901&23.110&0.164&\textbf{0.915}&\textbf{24.190}\\

\hline
\end{tabular}
}
\label{tab:comparison}

\end{table}

\subsection{Qualitative Evaluation}

As demonstrated in Figure~\ref{fig:comparison1}, our method demonstrates exceptional prowess in \textbf{local editing}, enabling precise modifications within specific regions of a 3D scene without affecting the overall integrity. Our method also excels in\textbf{ multi-attribute editing}, seamlessly combining changes across color, texture, and geometry within a single coherent edit. We also evaluate our method for the \textbf{object removal task}. The goal of 3D object removal is to eliminate an object from a 3D environment, potentially leaving behind voids due to the lack of observational data. For the object removal task, we identify and remove the regions based on the 2D mask, $M$. Subsequently, we focus on inpainting these "invisible regions" in the original 2D frames using LAMA ~\cite{suvorov2022resolution}. In Figure~\ref{fig:comparison-removal}, we demonstrate our 3DEgo's effectiveness in object removal compared to Gaussian Grouping. Our method's reconstruction output notably surpasses that of Gaussian Grouping~\cite{ye2023gaussian} in terms of retaining spatial accuracy and ensuring consistency across multiple views.

 \begin{wrapfigure}{r}{0.5\textwidth} 
    \centering
    
    \includegraphics[width=0.48\textwidth, trim={4cm 2.5cm 4cm 3cm}, clip]{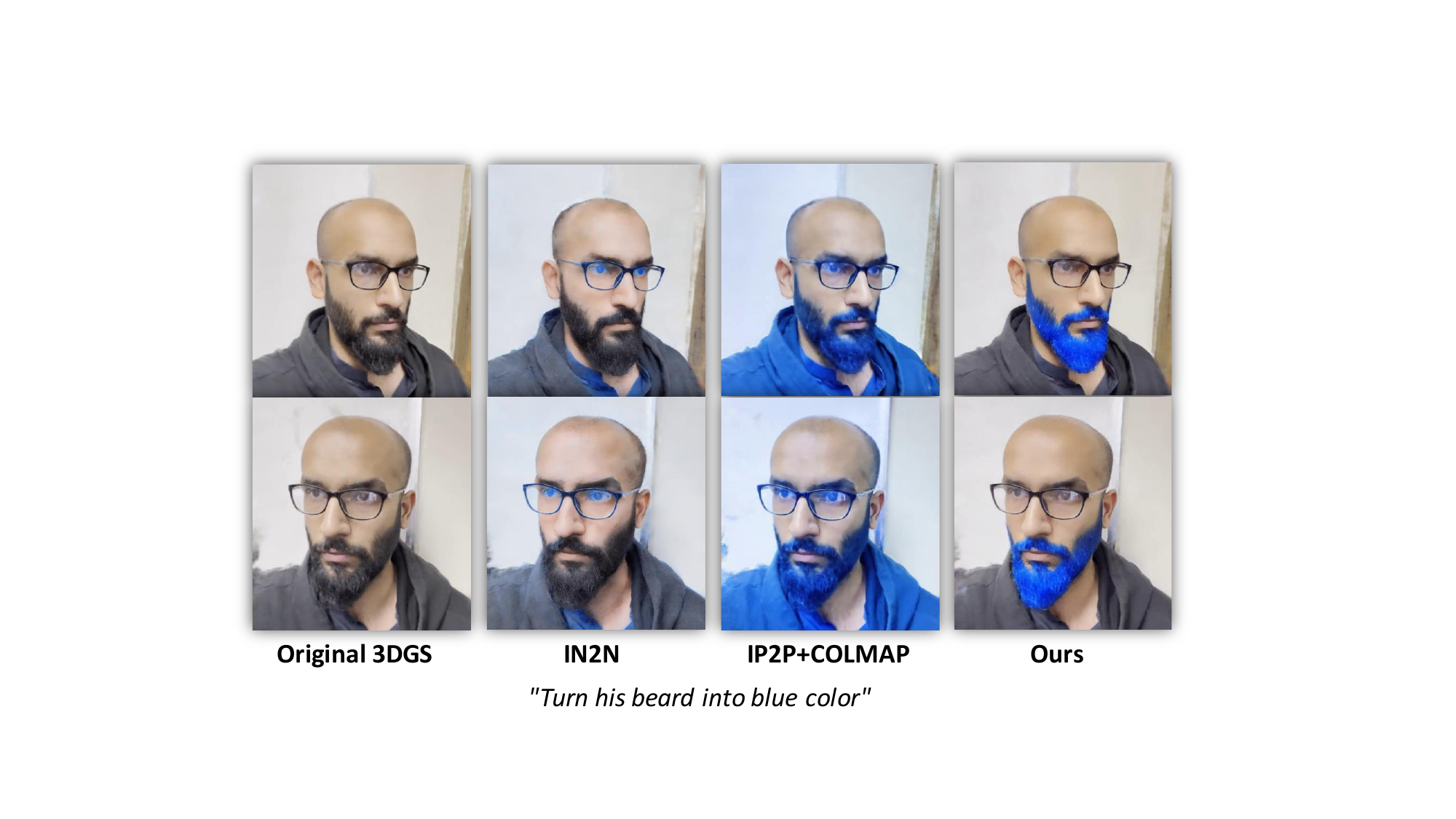} 
\caption{\footnotesize{ Our method, \textbf{3DEgo} achieves precise editing without using any SfM poses. To construct the IP2P+COLMAP 3D scene, we train nerfacto~\cite{tancik2023nerfstudio} model on IP2P~\cite{brooks2023instructpix2pix} edited frames.}}
\label{fig:fig2}
 
\end{wrapfigure}

\subsection{Quantitative Evaluation}

In our quantitative analysis, we employ three key metrics: {CLIP Text-Image Direction Similarity (CTIS)}~\cite{gal2021stylegan}, {CLIP Direction Consistency Score 
 (CDCR)}~\cite{haque2023instruct}, and {Edit PSNR (E-PSNR)}.
 We perform 200 edits across the six datasets listed above. We present quantitative comparisons with COLMAP-based 3D editing techniques in Table~\ref{tab:comparison}. Additionally, we extend our evaluation by integrating pose-free 3D reconstruction methods into our pipeline, with the performance outcomes detailed in Table~\ref{tab:comparison2}.

%% file: sec/06_limitations.tex

%% file: sec/06_conclusion.tex
\begin{table}[t]
\centering
\footnotesize
\setlength{\tabcolsep}{2pt} 
\renewcommand{\arraystretch}{1} 
\caption{\footnotesize\textbf{Comparing With Pose-Unknown Methods.} Quantitative analysis of 200 edits applied to six datasets, comparing methods proposed for NeRF reconstruction without known camera poses. The top-performing results are emphasized in bold.}
\resizebox{\textwidth}{!}{
\begin{tabular}{l|ccc|ccc|ccc}
\hline
\multirow{2}{*}{\textbf{Datasets}} & \multicolumn{3}{c|}{\textbf{BARF~\cite{lin2021barf}}} & \multicolumn{3}{c|}{\textbf{Nope-NeRF~\cite{bian2023nope}}} & \multicolumn{3}{c}{\textbf{Ours}} \\ 
& CTIS$\uparrow$ & CDCR$\uparrow$ & E-PSNR$\uparrow$ & CTIS$\uparrow$ & CDCR$\uparrow$ & E-PSNR$\uparrow$ & CTIS$\uparrow$ & CDCR$\uparrow$ & E-PSNR$\uparrow$ \\ 
\hline

GS25 (Ours) &0.139&0.797&20.478&0.128&0.753&19.660&\textbf{0.169}&\textbf{0.925}&\textbf{23.660}\\
Mip-NeRF &0.134&0.806&21.332&0.147&0.820&18.799&\textbf{0.175}&\textbf{0.901}&\textbf{24.250}\\
NeRFstudio &0.140&0.813&20.116&0.138&0.773&21.360&\textbf{0.163}&\textbf{0.931}&\textbf{24.990}\\
CO3D-V2 &0.157&0.820&21.148&0.129&0.824&17.971&\textbf{0.179}&\textbf{0.936}&\textbf{26.020}\\
IN2N &0.150&0.829&22.092&0.161&0.818&22.604&\textbf{0.183}&\textbf{0.925}&\textbf{26.390}\\
Tanks \& Temples &0.135&0.806&21.573&0.157&0.810&20.904&\textbf{0.164}&\textbf{0.915}&\textbf{24.190}\\

\hline
\end{tabular}
}
\label{tab:comparison2}
\end{table}

\section{Ablations}
To assess the influence of different elements within our framework, we employ PSNR, SSIM, and LPIPS metrics across several configurations. Given that images undergo editing before the training of a 3D model, our focus is on determining the effect of various losses on the model's rendering quality. The outcomes are documented in Table~\ref{table:ablation}, showcasing IP2P+COLMAP as the baseline, where images are edited using the standard IP2P approach~\cite{brooks2023instructpix2pix} and COLMAP-derived poses are utilized for 3D scene construction. 

\begin{wraptable}{r
}{0.5\textwidth} 

\centering
\caption{\footnotesize Ablation study results on \textbf{GS25} dataset.}
\setlength\tabcolsep{3pt}
\resizebox{0.95\linewidth}{!}{
  \begin{tabular}{lccc}
  \toprule
  Method & PSNR$\uparrow$ & SSIM$\uparrow$ & LPIPS$\downarrow$ \\
  \midrule
  Ours & \textbf{27.86} & \textbf{0.90} & \textbf{0.18} \\
  IP2P+COLMAP&23.87&0.79&0.23\\
  Ours w/o $L_{KEA}$ & 26.73 & 0.88 & 0.19
   \\
  Ours w/o $L_{ipc}$ & 22.46 & 0.0.78 & 0.24 \\
  Ours w/o $L_{pc}$ & 25.18 & 0.84 & 0.20 \\
  \bottomrule
  \end{tabular}%
 }

\label{table:ablation}
\end{wraptable}

Although the IP2P+COLMAP setup demonstrates limited textual fidelity due to editing inconsistencies (see Figure~\ref{fig:fig2}), we are only interested in the rendering quality in this analysis to ascertain our approach's effectiveness. Table~\ref{table:ablation} illustrates the effects of different optimization hyperparameters on the global scene expansion. The findings reveal that excluding $\mathcal{L}_{KEA}$ in the scene expansion process minimally affects rendering quality. On the other hand, omitting $\mathcal{L}_{ipc}$ leads to unwanted densification resulting in the inferior performance of the trained model. 
\begin{wrapfigure}{r}{0.50\textwidth} 
    \centering
    \includegraphics[width=0.50\textwidth, trim={1cm 7cm 12cm 7.5cm}, clip]{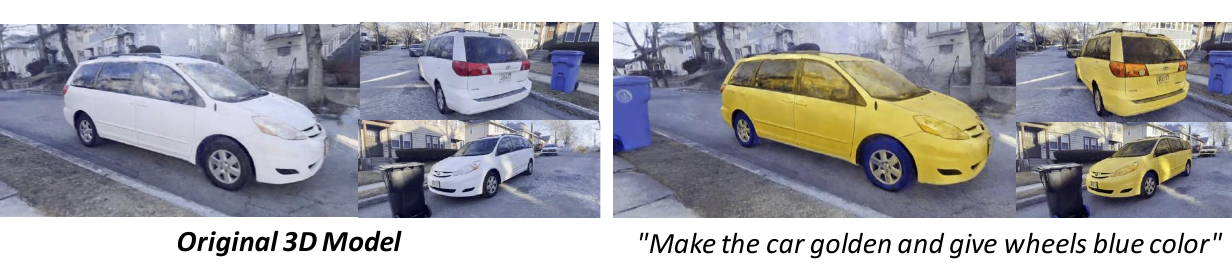} 
\caption{\footnotesize{Due to the limitations of the IP2P model, our method inadvertently alters the colors of the van's windows, which is not the desired outcome.}}
\label{fig:limitations}
\end{wrapfigure}
\section{Limitation}

Our approach depends on the pre-trained IP2P model~\cite{brooks2023instructpix2pix}, which has inherent limitations, especially evident in specific scenarios. For instance, Figure~\ref{fig:limitations} shows the challenge with the prompt \textit{``Make the car golden and give wheels blue color"}. Unlike IN2N~\cite{haque2023instruct}, which introduces unspecific color changes on the van's windows. Our method offers more targeted editing but falls short of generating ideal results due to IP2P's limitations in handling precise editing tasks.

\section{Conclusion}
\label{sec:conclusion}

\textit{3DEgo} marks a pivotal advancement in 3D scene reconstruction from monocular videos, eliminating the need for conventional pose estimation methods and model initialization. Our method integrates frame-by-frame editing with advanced consistency techniques to efficiently generate photorealistic 3D scenes directly from textual prompts. Demonstrated across multiple datasets, our approach showcases superior editing speed, precision, and flexibility. \textit{3DEgo} not only simplifies the 3D editing process but also broadens the scope for creative content generation from readily available video sources. This work lays the groundwork for future innovations in accessible and intuitive 3D content creation tools.